\documentclass{article} 
\usepackage[T1]{fontenc}
\usepackage{nips13submit_e,times}
\usepackage{hyperref}
\usepackage{url}
\usepackage[numbers]{natbib}
\usepackage{algorithm}
\usepackage{algorithmic}
\usepackage[pdftex]{graphicx}
\usepackage{subfigure} 
\usepackage{multirow}
\usepackage{multicol}
\usepackage{times}
\usepackage[small,it]{caption}
\usepackage{enumitem}
\usepackage{xcolor}
\usepackage{verbatim}
\usepackage{hyperref}
\usepackage{caption}
\usepackage{array}
\usepackage{relsize}
\usepackage[psamsfonts]{amssymb}
\usepackage{amsmath}
\usepackage{url}

\usepackage{rotating}
\usepackage{wrapfig,booktabs,blindtext}
\usepackage{color}
\newcommand{\todo}[1]{\textcolor{blue}{\textbf{#1}}}
\graphicspath{{fig/}}
\newcommand{\header}[1]{\noindent\textbf{#1}}

 \belowdisplayskip \abovedisplayskip

 \newlength\savedwidth

\newlength{\sectionReduceTop}
\newlength{\sectionReduceBot}
\newlength{\subsectionReduceTop}
\newlength{\subsectionReduceBot}
\newlength{\abstractReduceTop}
\newlength{\abstractReduceBot}
\newlength{\captionReduceTop}
\newlength{\captionReduceBot}
\newlength{\subsubsectionReduceTop}
\newlength{\subsubsectionReduceBot}

\newlength{\horSkip}
\newlength{\verSkip}

\newlength{\figureHeight}
\title{Learning Trajectory Preferences for Manipulators\\ via Iterative Improvement}

\author{
Ashesh Jain, Brian Wojcik, Thorsten Joachims, Ashutosh Saxena\\
Department of Computer Science, Cornell University.\\
\texttt{\{ashesh,bmw75,tj,asaxena\}@cs.cornell.edu}
}

\nipsfinalcopy 

\begin{document}

\maketitle


\vspace*{\abstractReduceTop}
\begin{abstract}
\vspace*{\abstractReduceBot}

We consider the problem of learning good trajectories for manipulation tasks.
This is challenging because the criterion defining a good trajectory varies with 
users, tasks and environments.  In this paper, we propose a co-active online 
learning framework for teaching robots the preferences of its users for object
manipulation tasks. The key novelty of our approach lies in the type of feedback
expected from the user: the human user does not need to demonstrate optimal trajectories 
as training data, but merely needs to iteratively provide trajectories that slightly
improve over the trajectory currently proposed by the system. We argue that this 
co-active preference feedback can be more easily elicited from the user than
demonstrations of optimal trajectories, which are often challenging and non-intuitive
to provide on high degrees of freedom manipulators. Nevertheless, theoretical
regret bounds of our algorithm match the asymptotic rates of optimal trajectory
algorithms. We demonstrate the generalizability of our algorithm on a variety of grocery checkout 
tasks, for whom, the preferences were not only influenced by the object being manipulated but also 
by the surrounding environment.\footnote{For more details and a demonstration video, visit: \url{http://pr.cs.cornell.edu/coactive}}

\end{abstract}


\vspace*{\sectionReduceTop}
\section{Introduction}
\vspace*{\sectionReduceBot}

Mobile manipulator robots have arms with high degrees of freedom (DoF), enabling them to perform household chores (e.g., PR2) or complex assembly-line tasks (e.g., Baxter). In performing these tasks, a key problem lies in identifying appropriate trajectories. An appropriate trajectory not only needs to be valid from a geometric standpoint (i.e., feasible and obstacle-free, the criterion that most path planners focus on), but it also needs to satisfy the user's preferences.

Such user's preferences over trajectories vary between users, between tasks, and between the environments the 
trajectory is performed in. For example, a household robot should move a glass of water 
in an upright position without jerks while maintaining a safe distance from nearby electronic devices.
In another example, a robot checking out a kitchen knife
at a grocery 
store should strictly move it at a safe distance from nearby humans. Furthermore, straight-line trajectories 
in Euclidean space may no longer be the preferred ones. For example, trajectories of
heavy items should not pass over fragile items but rather move around them. These preferences are 
often hard to describe and anticipate without knowing where and how the robot is deployed.
This makes it infeasible to manually encode (e.g.~\citep{Koppula13}) them in existing path planners (such as \cite{Chomp,Ompl}) a priori.

In this work we propose an algorithm for learning user preferences over trajectories through interactive feedback from the user in a co-active learning setting~\citep{Shivaswamy12}. Unlike in other learning settings, where a human first demonstrates optimal trajectories for a task to the robot, our learning model does not rely on the user's ability to demonstrate optimal trajectories a priori. Instead, our learning algorithm explicitly guides the learning process and merely requires the user to incrementally improve the robot's trajectories. From these interactive improvements the robot learns a general model of the user's preferences in an online fashion. We show empirically that a small number of such interactions is sufficient to adapt a robot to a changed task. \textit{Since the user does not have to demonstrate a (near) optimal trajectory to the robot},
we argue that our feedback is easier to provide and more widely applicable. Nevertheless, we will show that it leads 
to an online learning algorithm with provable regret bounds that decay at the same rate as if optimal 
demonstrations were available. 

In our empirical evaluation, we learn preferences for a high DoF Baxter robot on a
variety of grocery checkout tasks.
By designing expressive trajectory features, we show how our algorithm learns  
preferences from online user feedback on a broad range of tasks for which object properties are of 
particular importance (e.g., manipulating sharp objects with humans in vicinity). 
We extensively evaluate our approach on a set of 16 grocery checkout tasks, both
in batch experiments as well as through robotic experiments wherein users provide
their preferences on the robot.  Our results show that
robot trained using our algorithm not only quickly learns good trajectories on individual tasks, but also generalizes well to tasks that it has not seen before.
\vspace*{\sectionReduceTop}
\section{Related Work}
\label{relwork}
\vspace*{\sectionReduceBot}

Teaching a robot to produce desired motions has been a long standing goal 
and several approaches have been studied.  Most of the past research has focused on 
mimicking expert's demonstrations, for example,
autonomous helicopter flights~\citep{Abbeel10}, ball-in-a-cup experiment~\citep{Kober11}, 
planning 2-D paths~\citep{Ratliff07b,Ratliff09b,Ratliff06}, etc.  Such a setting (learning from demonstration, LfD) 
is applicable to scenarios when it is clear to an expert what constitutes a good trajectory.
In many scenarios, especially involving high DoF manipulators, this is extremely challenging to
do~\citep{Akgun12}.\footnote{Consider the following analogy:  In search engine results, it is much harder
for a user to provide the best web-pages for each query, but it is easier to provide relative ranking 
on the search results by clicking.}
 This is because the users have to give not only the end-effector's location at each time-step, 
but also the full configuration of the arm in a way that is spatially and temporally consistent.
In our setting, the user never discloses the optimal trajectory (or provide optimal feedback) 
to the robot, but instead, the robot learns preferences from sub-optimal suggestions on 
how the trajectory can be improved.

Some later works in LfD provided ways for handling noisy demonstrations, under the assumption that
demonstrations are either near optimal~\citep{Ziebart08} or locally optimal~\citep{Levine12}.  
Providing noisy demonstrations is different from providing relative preferences, which are biased and can be far from optimal. 
We compare with an algorithm for noisy LfD learning in our experiments.
A recent work~\citep{Wilson12} leverages user feedback to learn rewards of a Markov decision process. 
Our approach advances over~\citep{Wilson12} and Calinon et.~al.~\citep{Calinon07} in that it models sub-optimality
in user feedback and theoretically converges to user's hidden score function. We also capture
the necessary contextual information for household and assembly-line robots, while such context is absent in~\citep{Calinon07,Wilson12}. 
Our application scenario of learning trajectories for high DoF manipulations performing 
tasks in presence of different objects and environmental constraints goes beyond the
application scenarios that previous works have considered. 
We design appropriate features that consider
robot configurations, object-object relations, and temporal behavior, and use
them to learn a score function representing the preferences in trajectories.

User preferences have been studied in the field of human-robot
interaction. Sisbot et.~al.~\citep{Sisbot07,Sisbot07b} and Mainprice et.~al.~\citep{Mainprice11}
planned trajectories satisfying user specified preferences in form of constraints on the distance of robot from user, 
the visibility of robot and the user arm comfort. Dragan et.~al.~\citep{Dragan13b} used functional gradients~\citep{Chomp} 
to optimize for legibility of robot trajectories. We differ from these in that we \textit{learn} score functions reflecting user preferences from implicit feedback. 

\begin{figure}[t]
\centering
\vspace*{-0.07in}
\includegraphics[width=0.4\textwidth]{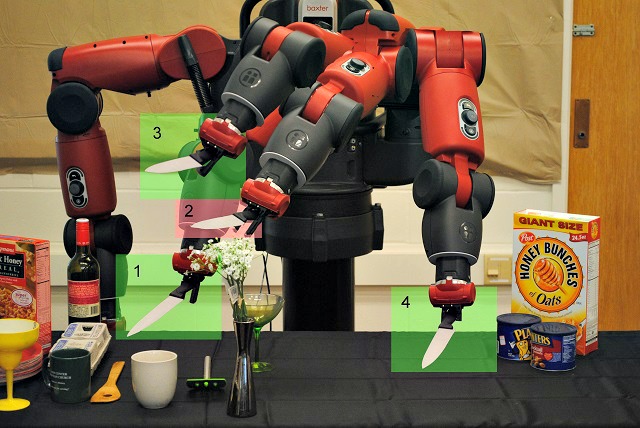}
\hspace*{5mm}
\includegraphics[width=0.4\textwidth]{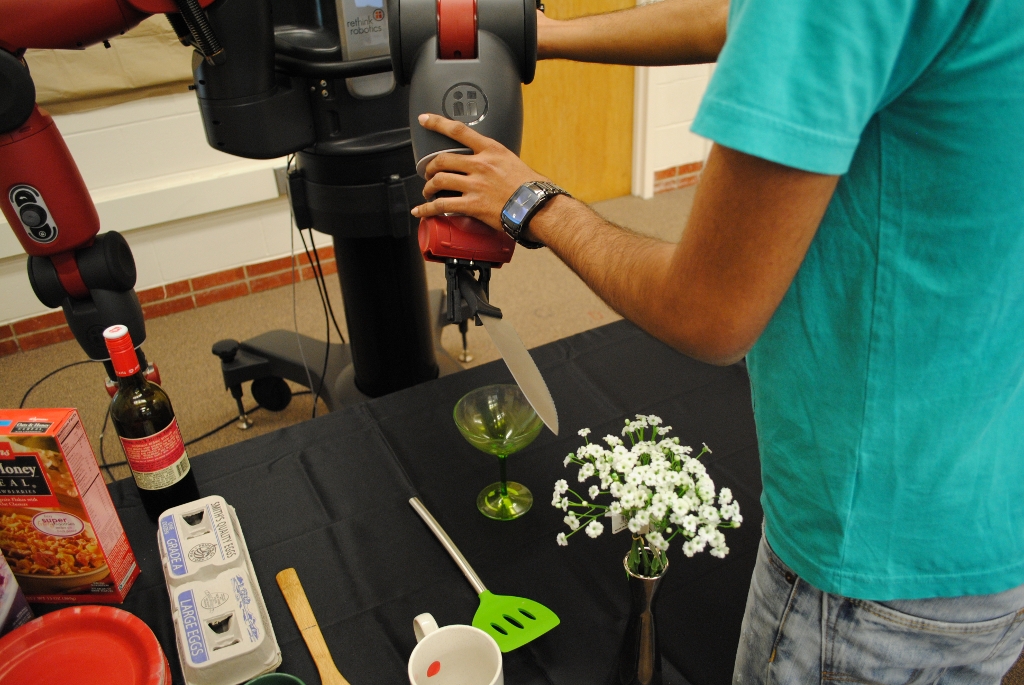}
\vspace*{\captionReduceTop}
\caption{\textbf{Zero-G feedback:} Learning trajectory preferences from sub-optimal
\emph{zero-G} feedback. (\textbf{Left}) 
Robot plans a bad trajectory (waypoints 1-2-4) with knife close to flower. As feedback, user corrects waypoint 2 and moves it to waypoint 3. (\textbf{Right}) User providing \emph{zero-G} feedback
on waypoint 2.}
\label{fig:trajvisualization}
\vspace*{\captionReduceBot}
\vspace*{\captionReduceBot}
\end{figure}

\vspace*{\sectionReduceTop}
\section{Learning and Feedback Model} \label{sec:coactmodel}
\vspace*{\sectionReduceBot}

We model the learning problem in the following way. For a given task, the robot is given a context $x$ 
that describes the environment, the objects, and any other input relevant to the
problem.  The robot has to figure out what is a good trajectory $y$ for this context.
Formally, we assume that the user has a scoring function $s^*(x,y)$ that reflects how much 
he values each trajectory $y$ for context $x$. The higher the score, the better the trajectory. 
Note that this scoring function cannot be 
observed directly, nor do we assume that the user can actually provide cardinal valuations 
according to this function. Instead, we merely assume that the user can provide us with 
{\em preferences} that reflect this scoring function. The robots goal is to learn a function $s(x,y;w)$ 
(where $w$ are the parameters to be learned) that approximates the users true scoring function $s^*(x,y)$ as closely as possible.

\textbf{Interaction Model.} The learning process proceeds through the following repeated cycle of interactions between robot and user. 

\textbf{Step 1:} The robot receives a context $x$. It then uses a planner to sample a set of trajectories, 
and ranks them according to its current approximate scoring function $s(x,y;w)$.\\
\textbf{Step 2:} The user either lets the robot execute the top-ranked trajectory, or 
corrects the robot by providing an improved trajectory $\bar{y}$. This provides 
feedback indicating that $s^*(x,\bar{y}) > s^*(x,y)$.\\
\textbf{Step 3:}  The robot now updates the parameter $w$ of $s(x,y;w)$ based on this preference feedback and returns to step 1.

\textbf{Regret.} The robot's performance will be measured in terms of regret, 
$REG_T = \frac{1}{T} \sum_{t=1}^T [ s^*(x_t,{y}^*_t) - s^*(x_t,{y}_t) ]$,
which compares the robot's trajectory ${y}_t$ at each time step $t$ against the optimal 
trajectory ${y}_t^*$ maximizing the user's unknown scoring function $s^*(x,{y})$, 
${y}_t^* = argmax_{y} s^*(x_t,{y})$.
Note that the regret is expressed in terms of the user's true scoring function $s^*$, 
even though this function is \emph{never observed}. Regret characterizes 
the performance of the robot over its whole lifetime, therefore reflecting how well it 
performs \emph{throughout} the learning process.
As we will show in the following sections, we employ learning algorithms with theoretical 
bounds on the regret for scoring functions that are linear in their parameters, 
making only minimal assumptions about the difference in score between $s^*(x,\bar{y})$ 
and $s^*(x,y)$ in Step 2 of the learning process.

\textbf{User Feedback and Trajectory Visualization.\label{subsec:tajgen}} 
Since the ability to easily give preference feedback in Step 2 is crucial for making 
the robot learning system easy to use for humans, we designed two feedback mechanisms 
that enable the user to easily provide improved trajectories.\\
\textit{(a) Re-ranking:} We rank trajectories in order of their current predicted scores and visualize the ranking
using OpenRave~\citep{Diankov10}. User observers trajectories sequentially 
and clicks on the first trajectory which is better than the top ranked trajectory.\\
\textit{(b) Zero-G:} This feedback allow users to improve trajectory waypoints by physically
changing the robot's arm configuration as shown in Figure~\ref{fig:trajvisualization}. 
To enable effortless steering of robot's arm to desired configuration we leverage Baxter's
zero-force gravity-compensation mode. Hence we refer this feedback as \emph{zero-G}.
This feedback is useful (i) for bootstrapping the robot,
(ii) for avoiding local maxima where the top trajectories in the ranked list are
all bad but ordered correctly, and (iii) when the user is satisfied with the top
ranked trajectory except for minor errors. A counterpart of this feedback is keyframe based
LfD~\citep{Akgun12} where an expert demonstrates a sequence of optimal waypoints instead of the complete trajectory.

Note that in both re-ranking and zero-G 
feedback, the user never reveals the optimal trajectory 
to the algorithm but just provides a slightly improved trajectory. 

\vspace*{0.4\sectionReduceTop}
\vspace*{\sectionReduceTop}
\section{Learning Algorithm}
\vspace*{\sectionReduceBot}
\label{sec:learningalgo}

For each task, we model the user's scoring function $s^*(x,y)$ with
the following parameterized family of functions. 
\vspace*{-1mm}
\begin{equation}
s(x,y;w) = w \cdot \phi(x,y)
\vspace*{-1mm}
\end{equation}
$w$ is a weight vector that needs to be learned, and $\phi(\cdot)$ are features
describing trajectory $y$ for context $x$. We further decompose the score function in
two parts, one only concerned with the objects the trajectory is interacting with, and the other with the object being 
manipulated and the environment.
\begin{align}
s(x,y;w_O,w_E) = s_O(x,y;w_O) + s_E(x,y;w_E)
= w_O \cdot \phi_O(x,y) + w_E \cdot \phi_E(x,y)
\label{eq:scorefn}
\end{align}


We now describe the features for the two terms, $\phi_O(\cdot)$ and $\phi_E(\cdot)$ in the following.

\vspace*{\subsectionReduceTop}
\subsection{Features Describing Object-Object Interactions}
\vspace*{\subsectionReduceBot}

\begin{figure}[t]
\centering
\includegraphics[width=\linewidth, height=0.24\linewidth]{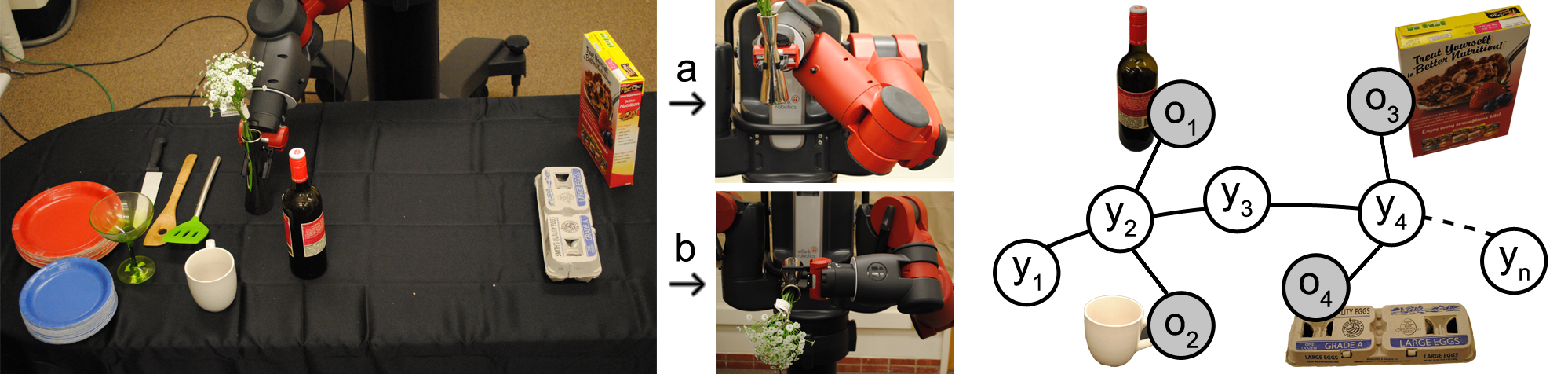}
\vspace*{\captionReduceTop}
\vspace*{\captionReduceTop}
\caption{\textbf{(Left)} A grocery checkout environment with a few objects where the robot was asked to checkout flowervase on the left to the right. \textbf{(Middle)}
There are two ways of moving it, `a' and `b', both are sub-optimal in that the arm 
is contorted in `a' but it tilts the vase in `b'. Given such constrained scenarios, we need to reason about such subtle preferences. \textbf{(Right)} We encode 
preferences concerned with object-object interactions in a score function expressed over a graph. Here $y_1,\ldots,y_n$ are different waypoints in a trajectory.
The shaded nodes corresponds to environment (table node not shown here).
Edges denotes interaction between nodes.}
\vspace*{\captionReduceBot}
\vspace*{\captionReduceBot}
\label{fig:trajgraph}
\end{figure}

This feature captures the interaction between
objects in the environment with the object being manipulated. 
We enumerate waypoints of trajectory $y$ as $y_1,..,y_N$ and
objects in the environment as $\mathcal{O}=\{o_1,..,o_K\}$. The robot manipulates the
object $\bar{o}\in \mathcal{O}$. 
A few of the trajectory waypoints would be affected by the other objects in the environment.
For example in Figure~\ref{fig:trajgraph}, $o_1$ and $o_2$ affect 
the waypoint $y_3$ because of proximity.
Specifically, we connect an object $o_k$ to a trajectory waypoint if
the minimum distance to collision is less than a threshold or if $o_k$
lies below $\bar{o}$.  The edge connecting $y_j$ and
$o_k$ is denoted as $(y_j,o_k)\in\mathcal{E}$. 

Since it is the attributes \cite{Koppula11} of the object that really matter in 
determining the trajectory quality, we represent each object with its \emph{attributes}.
Specifically, for every object $o_k$, we consider a vector of $M$ binary
variables [$l_k^1,..,l_k^M$], with each $l_k^m = \{0,1\}$ indicating whether
object $o_k$ possesses property $m$ or not.
  For example, if the set of possible properties are 
\{heavy, fragile, sharp, hot, liquid, electronic\}, then a laptop and a glass table can have 
labels [$0,1,0,0,0,1$] and [$0,1,0,0,0,0$] respectively. 
The
binary variables $l_k^p$ and $l^q$ indicates whether $o_k$ and $\bar{o}$ possess property $p$ and $q$ 
respectively.\footnote{In this work, our goal is to relax the assumption of unbiased 
and close to optimal feedback. We therefore assume complete knowledge of the environment for our algorithm, 
and for the algorithms we compare against. In practice, such knowledge can be extracted using an object attribute
labeling algorithm such as in \cite{Koppula11}.} Then, for every ($y_j,o_k$) edge, we extract following four features 
$\phi_{oo}(y_j, o_k)$: projection of minimum distance to collision along x, y and z (vertical) axis and a binary
variable, that is 1, if $o_k$ lies vertically below $\bar{o}$, 
0 otherwise. 

We now define the score $s_O(\cdot)$ over this graph as follows:
\begin{align}
s_{O}(x,y;w_{O}) =   \sum_{(y_j,o_k)\in\mathcal{E}} \sum_{p,q=1}^M l_k^p l^q
[w_{pq}\cdot\phi_{oo}(y_j, o_k)]
\end{align}
Here, the weight vector
$w_{pq}$ captures interaction between objects with properties $p$ and $q$. 
We obtain $w_O$ in eq.~\eqref{eq:scorefn} by concatenating vectors $w_{pq}$. 
More formally, if the vector at position $i$
of $w_{O}$ is $w_{uv}$ then the vector corresponding to position $i$ of $\phi_O(x,y)$
will be $\sum_{(y_j,o_k)\in\mathcal{E}} l_k^u l^v
[\phi_{oo}(y_j, o_k)]$. 

\vspace*{\subsectionReduceTop}
\subsection{Trajectory Features}
\vspace*{\subsectionReduceBot}

We now describe features, $\phi_{E}(x,y)$, obtained by performing operations on a set of waypoints. 
They comprise the following three types of the features:

\header{Robot Arm Configurations.}
While a robot can reach the same operational
space configuration for its wrist with different configurations of the arm, not all
of them are preferred~\cite{Zacharias11}. For example, the contorted way of holding
the flowervase shown in Figure~\ref{fig:trajgraph}
may be fine at that time instant, but would present problems if our goal is to 
perform an activity with it, e.g. packing it after checkout. Furthermore,
 humans like to anticipate robots move and to gain users' confidence, 
robot should produce predictable and legible robot motion~\citep{Dragan13b}. 

We compute features capturing robot's arm configuration using the location of its
elbow and wrist, w.r.t.\ to its shoulder, in cylindrical  coordinate system, 
$(r, \theta, z)$. We divide a trajectory into three parts in time and compute 9 
features for each of the parts. These features encode the maximum and minimum 
$r$, $\theta$ and $z$ values for wrist and elbow in that part of the trajectory, 
giving us 6 features. Since at the limits of the manipulator configuration, joint 
locks may happen, therefore we also add 3 features for the location of robot's 
elbow whenever the end-effector attains its maximum
$r$, $\theta$ and $z$ values respectively.
Therefore obtaining $\phi_{robot}(\cdot) \in \mathbb{R}^{9}$ (3+3+3=9) features for each one-third part 
and $\phi_{robot}(\cdot) \in \mathbb{R}^{27}$ for the complete trajectory.
\begin{wrapfigure}{0}{0.4\textwidth}
 \vskip -.3in
\centering
\includegraphics[width=0.7\linewidth]{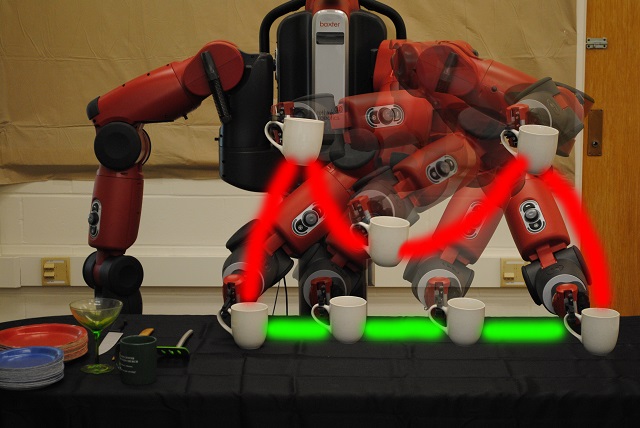}
\includegraphics[width=0.48\linewidth,height=0.7in,clip=true,trim=10 0 20 0]{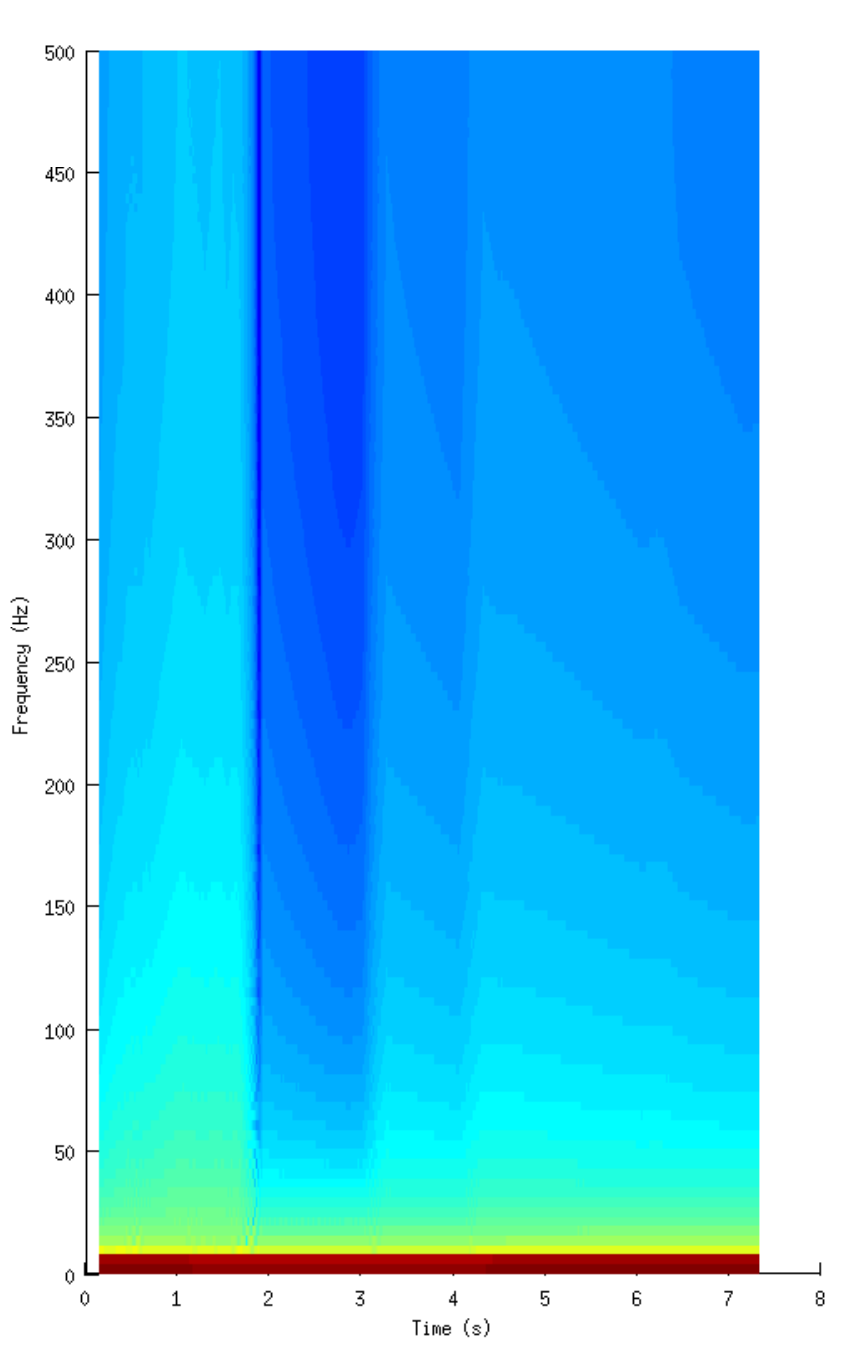}
\includegraphics[width=0.48\linewidth,height=0.7in,clip=true,trim=10 0 20 0]{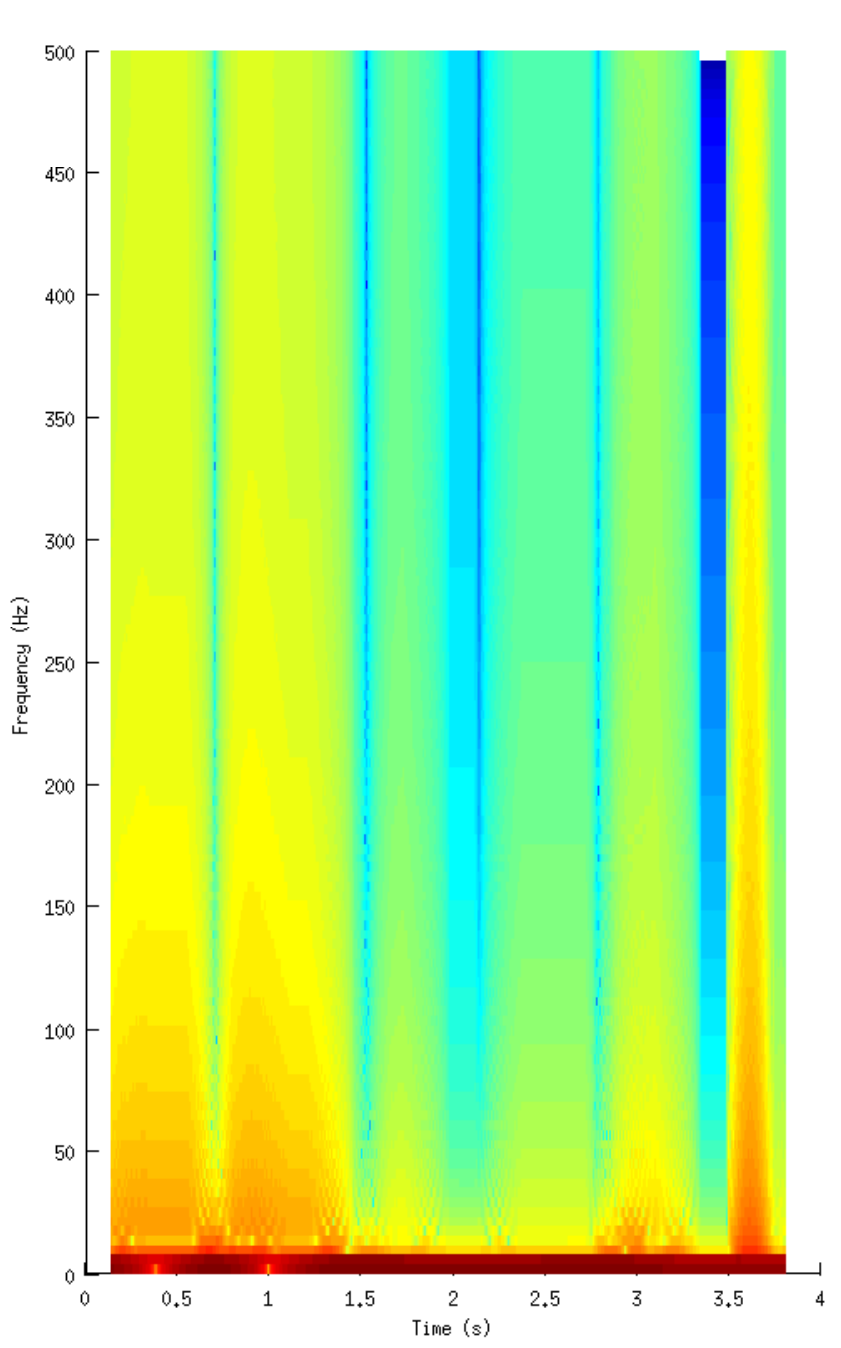}
\vspace*{\captionReduceTop}
\caption{\textbf{(Top)} A {\color{green}good} and {\color{red}bad} trajectory for moving a 
mug. The bad trajectory undergoes ups-and-downs. \textbf{(Bottom)} Spectrograms for movement 
in z-direction: \textbf{(Right)} {\color{green}Good} 
trajectory, \textbf{(Left)} {\color{red}Bad} trajectory. }
\vspace*{\captionReduceBot}
\label{fig:spect}
\end{wrapfigure}

\header{Orientation and Temporal Behavior of the Object to be Manipulated.}   
{Object orientation during the trajectory is crucial in deciding its quality.} 
For some tasks, the orientation
must be strictly maintained (e.g., moving a cup full of coffee); 
and for some others,
it may be necessary to  change it in a particular fashion (e.g., pouring activity).
Different parts of the trajectory 
may have different requirements over time.
For example, in the placing task, we may
need to bring the object closer to obstacles and be more careful. 

We therefore divide trajectory into three parts in time. 
For each part we store the cosine of the object's maximum 
deviation, along the vertical axis,  from  its final orientation at the goal location. To capture 
object's oscillation along trajectory,
we obtain a spectrogram for each one-third part for the movement of the object
in $x$, $y$, $z$ directions as well as for the deviation along vertical axis (e.g. Figure~\ref{fig:spect}).
We then compute the average power spectral density in the low and high frequency part as  eight 
additional features for each.
This gives us 9 (=1+4*2) features for each one-third part.  Together with one 
additional feature of object's maximum deviation along the whole trajectory,  
we get $\phi_{obj}(\cdot) \in \mathbb{R}^{28}$
(=9*3+1).

\header{Object-Environment Interactions.} 
This feature captures temporal variation of vertical and horizontal distances of the object
$\bar{o}$ from its surrounding surfaces.  In detail, we divide the trajectory into three equal
parts, and for each part we compute object's:
(i) minimum vertical distance from the nearest surface below it. 
(ii) minimum horizontal distance from the surrounding surfaces; and (iii)
minimum distance from the table, on which the task is being performed, and (iv) minimum distance
from the goal location. We also take an average, over all the waypoints, of the horizontal and vertical distances
between the object and the nearest surfaces around it.\footnote{We query PQP collision 
checker plugin of OpenRave for these distances.} 
To capture temporal variation 
of object's distance from its surrounding we plot a time-frequency spectrogram of the object's vertical distance from the nearest surface below it, from which we extract
six features by dividing it into grids. This feature is expressive enough to differentiate
whether an object just grazes over table's edge (steep change in vertical distance) 
versus, it first goes up and over the table and then moves down (relatively smoother change). 
Thus,  the features obtained from object-environment interaction are $\phi_{obj-env}(\cdot) \in 
\mathbb{R}^{20}$ (3*4+2+6=20).

Final feature vector is obtained by concatenating $\phi_{obj-env}$, $\phi_{obj}$ and $\phi_{robot}$,
giving us $\phi_E(\cdot) \in \mathbb{R}^{75}$.

\vspace*{0.7\subsectionReduceTop}
\vspace*{\subsectionReduceTop}
\vspace*{\subsectionReduceTop}
\subsection{Computing Trajectory Rankings\label{sec:plannerranking}}
\vspace*{\subsectionReduceBot}
For obtaining the top trajectory  (or a top few) for a given task with context $x$, we would like
to maximize the current scoring function $s(x,y;w_O,w_E)$.
\begin{equation}
y^*  = \arg\max_y s(x, {y};w_O,w_E). \label{eq:argmax_rank}
\end{equation}
Note that this poses two challenges. First, trajectory space is 
continuous and needs to be discretized to maintain argmax in~\eqref{eq:argmax_rank} tractable.
Second, for a given set $\{y^{(1)},\ldots,y^{(n)}\}$ of 
discrete trajectories, we need to compute ($\ref{eq:argmax_rank}$). 
Fortunately, the latter problem is easy to solve and simply amounts 
to sorting the trajectories by their trajectory scores $s(x,y^{(i)};w_O,w_E)$. 
Two effective ways of solving the former problem is either discretizing the robot's configuration space or directly
sampling trajectories from the continuous space. Previously both
approaches~\citep{Alterovitz07,Berg10,Dey12,Vernaza12} have been studied. 
However, for high DoF manipulators sampling based approaches~\citep{Berg10,Dey12} maintains
tractability of the problem, hence we take this approach. More precisely, similar to~\citet{Berg10},
we sample trajectories using rapidly-exploring random tree (RRT)~\citep{Lavalle01}.\footnote{When RRT 
becomes too slow, we switch to a more efficient bidirectional-RRT. 
The cost function (or its approximation) we learn can be fed to 
trajectory optimizers like CHOMP~\citep{Chomp} or optimal planners like RRT*~\citep{Karaman10} to produce 
reasonably good trajectories.}
Since our primary goal is to learn a score function on sampled set of
trajectories we now describe our learning algorithm and for more literature on sampling trajectories we
refer the readers to~\citep{Green07}.

\vspace*{\subsectionReduceTop}
\subsection{Learning the Scoring Function} 
\label{sec:learningscore}
\vspace*{\subsectionReduceBot}
The goal is to learn the parameters $w_O$ and $w_E$
of the scoring function $s(x,y;w_O,w_E)$ so that it can be used to rank trajectories according to the 
user's preferences. To do so, we adapt the Preference Perceptron algorithm \cite{Shivaswamy12} as 
detailed in Algorithm~\ref{alg:coactive}. We call this algorithm the Trajectory Preference Perceptron (TPP). 
Given a context $x_t$, the top-ranked trajectory $y_t$ under the current parameters $w_O$ and $w_E$, and 
the user's feedback trajectory $\bar{y}_t$, the TPP updates the weights in the direction 
${\phi}_O(x_t,\bar{y}_t) - {\phi}_O(x_t,{y}_t)$ and
${\phi}_E(x_t,\bar{y}_t) - {\phi}_E(x_t,{y_t})$ respectively. 

Despite its simplicity and even though the algorithm typically does not receive the optimal trajectory $y_t^* = \arg\max_ys^*(x_t,y)$ as feedback, the TPP enjoys guarantees on the regret \cite{Shivaswamy12}.
We merely need to characterize by how much the feedback improves on the presented 
ranking using the following definition of expected $\alpha$-informative feedback:
$
E_t[s^*(x_t,\bar{y}_t) ] \geq s^*(x_t,y_t) + 
\alpha ( s^*(x_t,y^*_t) - s^*(x_t,y_t)) - \xi_t
$.
This definition states that the user feedback should have a score of ${\bar{y}}_t$ that 
is---in expectation over the users choices---higher than that of ${y}_t$ by a fraction $\alpha
\in (0,1]$ of the maximum possible range $s^*(x_t,{\bar{y}}_t) - s^*(x_t,{y}_t)$. If
this condition is not fulfilled due to bias in the feedback, the slack variable $\xi_t$ captures the amount of violation. In this way any feedback can be described by an appropriate combination of $\alpha$ and $\xi_t$.
 Using these two parameters, 
the proof by \citep{Shivaswamy12} can be adapted to show that 
the expected average regret of the TPP is upper bounded by 
$
E[REG_T] \le \mathcal{O} (\frac{1}{\alpha\sqrt{T}} + \frac{1}{\alpha T}\sum_{t=1}^T \xi_t )
$
after $T$ rounds of feedback.
\begin{wrapfigure}{r}{0.54\textwidth}
\vskip -.17in
\begin{minipage}{0.54\textwidth}
  \begin{algorithm}[H]
\caption{Trajectory Preference Perceptron. (TPP) } 
\begin{algorithmic}
\STATE Initialize ${w}_O^{(1)} \leftarrow 0$, ${w}_E^{(1)} \leftarrow 0$
\FOR{$t = 1 \;\text{to}\; T$}
	\STATE Sample trajectories $\{y^{(1)},...,y^{(n)}\}$
	\STATE ${y}_t = argmax_{{y}} s(x_t,{y}; w_O^{(t)},w_E^{(t)})$
	\STATE Obtain user feedback $\bar{{y}}_t$
	\STATE $w_O^{(t+1)} \leftarrow w_O^{(t)} + {\phi}_O(x_t,\bar{{y}}_t) -
	{\phi}_O(x_t,{{y}}_t)$
	\STATE $w_E^{(t+1)} \leftarrow w_E^{(t)} + {\phi}_E(x_t,\bar{{y}}_t) - {\phi}_E(x_t,{{y}}_t)$

\ENDFOR
\end{algorithmic}
\label{alg:coactive}
\end{algorithm}
\end{minipage}
\vskip -.3in
\end{wrapfigure}

\vspace*{\sectionReduceTop}
\vspace*{\sectionReduceTop}
\vspace*{\sectionReduceTop}
\section{Experiments and Results}
\vspace*{\sectionReduceBot}
We now describe our data set, baseline algorithms and the evaluation metrics 
we use. Following this, we present quantitative results (Section~\ref{sec:results}) 
and report robotic experiments on Baxter (Section~\ref{sec:user-study}). 

\vspace*{\subsectionReduceTop}
\subsection{Experimental Setup}
\vspace*{\subsectionReduceBot}
\header{Task and Activity Set for Evaluation.}
We evaluate our approach on 16 pick-and-place robotic tasks in a grocery store checkout 
setting. To assess generalizability of our approach, for each task we train and test on scenarios 
with different objects being manipulated, and/or with a different environment.
We evaluate the quality of trajectories after the robot has grasped the items and 
while it moves them for checkout. Our work complements previous works on grasping items \citep{saxena2008roboticgrasping,lenz2013_deeplearning_roboticgrasp}, 
pick and place tasks~\citep{jiang2012placingobjects}, and
detecting bar code for grocery checkout~\citep{Klingbeil11}. We consider following three commonly occurring activities in a grocery store:
\begin{figure}[b]
\centering
\vspace*{\captionReduceTop}
\vspace*{\captionReduceTop}
\includegraphics[width=0.25\textwidth]{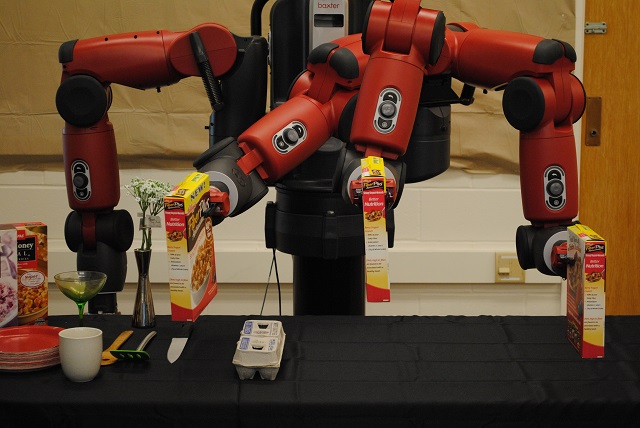}
\hskip 5mm
\includegraphics[width=0.25\textwidth]{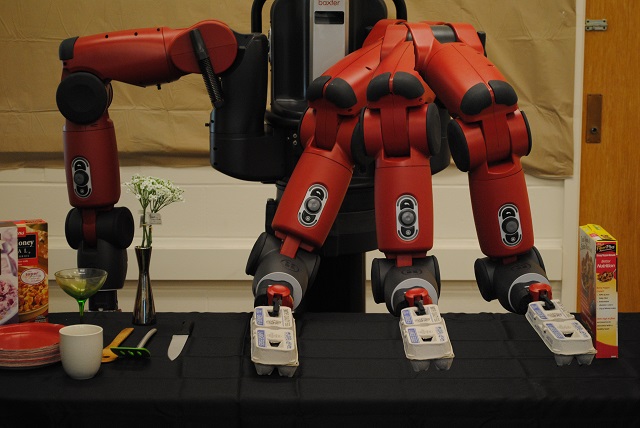}
\hskip 5mm
\includegraphics[width=0.25\textwidth]{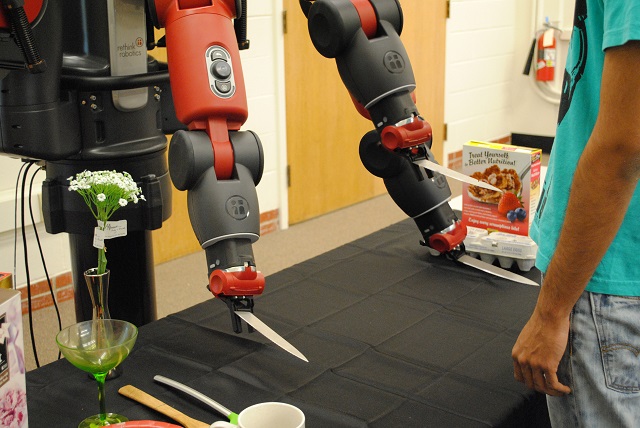}
\vspace*{\captionReduceTop}
\caption{(\textbf{Left}) \textit{Manipulation centric:} a box of 
cornflakes doesn't interact much with 
surrounding items and is indifferent to orientation. (\textbf{Middle}) 
\textit{Environment centric:} an egg carton is fragile and should preferably be kept upright and
closer to a supporting surface. (\textbf{Right}) \textit{Human centric:} a knife is sharp and 
interacts with nearby soft items and humans. It should strictly be kept at a safe distance from humans.}
\vspace*{\captionReduceBot}
\label{fig:taskcats}
\end{figure}

\noindent \emph{1) Manipulation centric:} These activities primarily care for the object 
being manipulated. 
Hence the object's properties and the way robot moves it in the environment 
is more relevant. Examples include moving common objects 
like cereal box, Figure~\ref{fig:taskcats} (left), or moving fruits and vegetables, which can be 
damaged when dropped/pushed into other items.\\
\emph{2) Environment centric:} These activities also care for the interactions of the object
being manipulated with the surrounding objects. Our object-object interaction
features allow the algorithm to learn preferences on trajectories for moving
fragile objects like glasses and egg cartons, Figure~\ref{fig:taskcats} (middle).\\
\emph{3) Human centric:} 
Sudden movements by the robot put the human in a danger of getting hurt. We consider activities 
where a robot manipulates sharp objects, e.g., moving a knife with a human in vicinity as shown 
in Figure~\ref{fig:taskcats} (right). In previous work, such relations were considered in the context
of scene understanding \cite{jiang2012humancontext,jiang-hallucinatinghumans-labeling3dscenes-cvpr2013}.

\header{Baseline algorithms.} 
We evaluate the algorithms that learn preferences from online feedback, under two settings:
(a) \emph{untrained}, where the algorithms learn preferences 
for the new task from scratch without observing any previous feedback; 
(b) \emph{pre-trained}, where the algorithms are pre-trained 
on other similar tasks, and then adapt 
to the new task.
We compare the following algorithms:
\begin{itemize}[leftmargin=*]
 \setlength{\itemsep}{-2pt}
 \setlength{\topsep}{-5pt}
  \setlength{\parsep}{0pt}
\item \emph{Geometric}: It plans a path, independent of the task, using a BiRRT~\citep{Lavalle01} planner.
\item \emph{Manual}:  It plans a path following certain manually coded preferences.
\item \emph{TPP}:  This is our algorithm. We evaluate it under both, \emph{untrained} and \emph{pre-trained} settings.
\item \emph{Oracle-svm}: This algorithm leverages the expert's labels on trajectories (hence the name \textit{Oracle}) and is trained 
using SVM-rank~\citep{Joachims06} in a batch manner. This algorithm is \emph{not realizable in practice}, as it requires labeling on the large space of trajectories.
We use this only in pre-trained setting and during prediction it just predicts once and does not learn further.
\item \emph{MMP-online}: This is an online implementation of Maximum margin planning (MMP)~\citep{Ratliff06,Ratliff09a}
algorithm. MMP attempts to make an expert's trajectory better than any other trajectory by a 
margin, and can be interpreted as a special case of our algorithm with
1-informative feedback. However, adapting MMP to our experiments poses two
challenges: (i) we do not have knowledge of optimal trajectory; and (ii) the state space 
of the manipulator we consider is too large, and discretizing makes learning via
MMP intractable. We therefore train MMP from online user feedback observed on a set of trajectories. 
We further treat the observed feedback as optimal. 
At every iteration we train a structural support vector machine
(SSVM)~\cite{Joachims09} using all previous feedback as training examples, 
and use the learned weights to predict trajectory scores for the next iteration. 
Since we learn on a set of trajectories, the argmax operation in 
SSVM remains tractable.  We quantify closeness of trajectories by the $l_2-$norm of difference in 
their feature representations, and choose the regularization
parameter $C$ for training SSVM in hindsight, to give an unfair advantage to MMP-online.
\end{itemize}
\vspace*{\subsectionReduceTop}
\header{Evaluation metrics.}
In addition to performing a user study on Baxter robot (Section~\ref{sec:user-study}), we also 
designed a data set to quantitatively evaluate the performance of our online algorithm. 
An expert labeled 1300 trajectories on a Likert scale of 1-5 (where 5 is the best)
on the basis of subjective 
human preferences.
Note that these absolute ratings are never provided
to our algorithms and are only used for the quantitative evaluation of different algorithms.
We quantify the quality of a ranked list of trajectories by its normalized
discounted cumulative gain (nDCG)~\citep{Manning08} at positions 1 and 3. 
While nDCG@1 is a suitable metric for autonomous robots that execute the top ranked
trajectory, nDCG@3 is suitable for scenarios where the robot is supervised
by humans.

\vspace*{\subsectionReduceTop}
\subsection{Results and Discussion}
\vspace*{\subsectionReduceBot}
\label{sec:results}

\noindent We now present the quantitative results on the data set of 1300 labeled trajectories. 

\begin{figure}[t]
\centering
\begin{tabular}{@{\hskip -.0in}l@{}c@{\hskip -.17in}c@{\hskip -.17in}c@{}}
\raisebox{.5in}{\rotatebox[origin=c]{90}{\scriptsize nDCG@3}}
&
\subfigure[\hspace*{-1mm}\scriptsize{Same environment, different object}]{
\includegraphics[width=0.34\linewidth,height=0.24\textwidth]{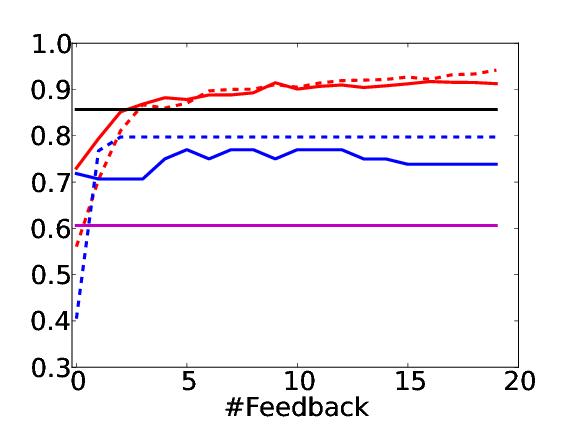}
\label{fig:genobj}
}
&
\subfigure[\hspace*{-1mm}\scriptsize{New Environment, same object}]{
\includegraphics[width=0.34\linewidth,height=0.24\textwidth]{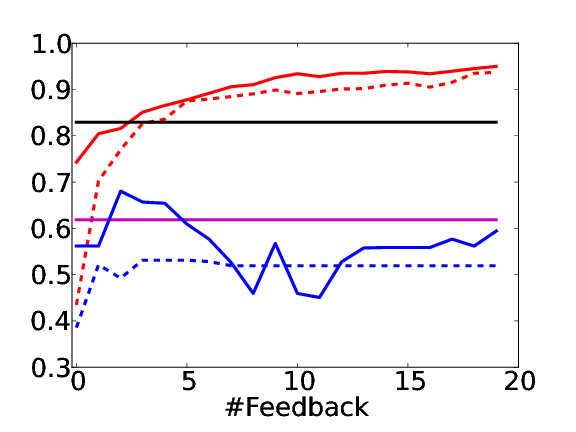}
\label{fig:genenv}
}
&
\subfigure[\hspace*{-1mm}\scriptsize{New Environment, different object}]{
\includegraphics[width=0.34\linewidth,height=0.24\textwidth]{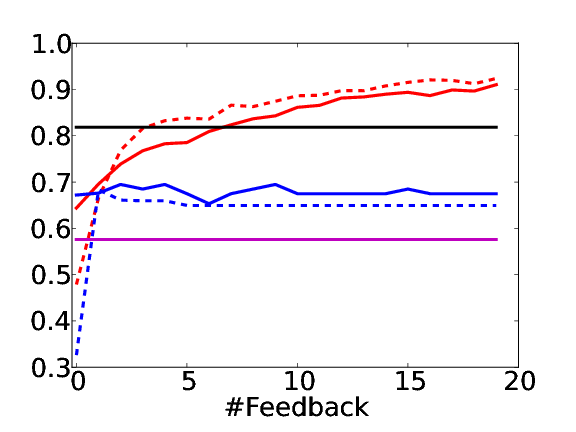}
\label{fig:genenvobj}
}
\end{tabular}
\vspace*{\captionReduceTop}
\vspace*{\captionReduceTop}
\caption{{Study of generalization with change in object, environment and both.} {\color{magenta}{Manual}}, Oracle-SVM, \textcolor{blue}{Pre-trained MMP-online (---)}, \textcolor{blue}{Untrained MMP-online (-- --)}, \textcolor{red}{ Pre-trained TPP (---)}, \textcolor{red}{ Untrained TPP (-- --)}. }
\vspace*{\captionReduceBot}
\vspace*{\captionReduceBot}
\label{fig:genstudy}
\end{figure}

\header{How well does TPP generalize to new tasks?} To study generalization of preference feedback
we evaluate performance of TPP-pre-trained (i.e., \textit{TPP} algorithm under \textit{pre-trained} setting)
 on a set of tasks the algorithm has not seen before.
We study generalization when: (a) only the object being
manipulated changes, e.g., an egg carton replaced by tomatoes, (b)
only the surrounding environment changes, e.g., rearranging objects in the
environment or changing the start location of tasks, and (c) when both change.
Figure~\ref{fig:genstudy} shows nDCG@3 plots averaged over tasks for all types of activities.\footnote{Similar 
results were obtained with nDCG@1 metric. We have not included it due to space constraints.}
TPP-pre-trained starts-off with higher nDCG@3 values than TPP-untrained in all three cases. 
Further, as more feedback is received, performance of both
algorithms improve to eventually become (almost) identical. 
We further observe, generalizing to tasks with both new environment and object is harder than when only 
one of them changes.

\begin{wraptable}{o}{0.58\linewidth}
\vspace*{\captionReduceTop}
\vspace*{\captionReduceTop}
\caption{\textbf{Comparison of different algorithms and study of features in untrained setting.} Table contains average nDCG@1(nDCG@3)
values over 20 rounds of feedback.}
\vspace*{\captionReduceBot}
 \resizebox{0.58\textwidth}{!}{
\centering
 \begin{tabular}{r@{}rrccc@{\hskip 0.1in}c}\cline{2-7}
 &&\multirow{2}{*}{Algorithms}&Manipulation &Environment&Human&\multirow{2}{*}{Mean}\\
 &&&centric&centric&centric&\\\cline{2-7}
 &&Geometric&0.46  (0.48)&0.45  (0.39)&0.31  (0.30)&0.40  (0.39)\\
 &&Manual&0.61 (0.62)&0.77  (0.77)&0.33  (0.31)&0.57  (0.57)\\\cline{2-7}
 \multirow{4}{*}{\raisebox{.08in}{\rotatebox[origin=c]{90}{TPP}}}&\multirow{4}{*}{\raisebox{.1in}{\rotatebox[origin=c]{90}{Features}}}&Obj-obj interaction&0.68 (0.68)&0.80 (0.79)&0.79 (0.73)&0.76 (0.74)\\
 &&Robot arm config&0.82 (0.77)&0.78 (0.72)&0.80 (0.69)&0.80 (0.73)\\
 &&Object trajectory&0.85 (0.81)&0.88 (0.84)&0.85 (0.72)&0.86 (0.79)\\
 &&Object environment&0.70 (0.69)&0.75 (0.74)&0.81 (0.65)&0.75 (0.69)\\\cline{2-7}
 &&TPP  (all features)&\textbf{0.88 (0.84)}&\textbf{0.90 (0.85)}&\textbf{0.90 (0.80)}&\textbf{0.89 (0.83)}\\
 &&MMP-online&0.47 (0.50)&0.54 (0.56)&0.33 (0.30)&0.45 (0.46)\\
 \cline{2-7}
 \end{tabular}
}
\vspace*{\captionReduceBot}
 \label{tab:algostudy}
 \end{wraptable}
\header{How does TPP compare to other algorithms?} Despite the fact that TPP never observes
 optimal feedback, it performs better than baseline algorithms, see Figure~\ref{fig:genstudy}. 
 It improves over Oracle-SVM in less than 5 feedbacks, which is not updated since it requires 
expert's labels on test set and hence it is impractical.
 MMP-online assumes every user feedback as optimal, and over iterations accumulates many contradictory training examples. 
This also highlights the sensitivity of MMP to sub-optimal demonstrations.  
We also compare against planners with manually coded preferences e.g., keep 
a flowervase upright. However, some preferences are difficult to specify, e.g., 
not to move heavy objects over fragile items. We empirically found  
the resulting manual algorithm produces poor trajectories with
an average nDCG@3 of 0.57 over all types of activities. 

\header{How helpful are different features?} Table~\ref{tab:algostudy} shows the performance of the TPP algorithm in 
the untrained setting using different features. Individually each feature captures several aspects indicating goodness of trajectories,
 and combined together they give the best performance.
Object trajectory features capture preferences related to the orientation of the object. 
Robot arm configuration and object environment features capture preferences 
by detecting undesirable contorted arm configurations and maintaining safe distance from surrounding surfaces, respectively. 
Object-object features by themselves can only learn, for example, to move egg carton closer to a supporting surface, 
but might still move it with jerks or contorted arms. These features can be combined with other features to yield more expressive features.
Nevertheless, by themselves they perform better than Manual algorithm. 
Table~\ref{tab:algostudy} also compares TPP and MMP-online under untrained setting.

\subsection{Robotic Experiment: User Study in learning trajectories}
\vspace*{\subsectionReduceBot}
\label{sec:user-study}
\noindent We perform a user study of our system on Baxter robot on a variety of tasks of varying difficulties.
Thereby, 
showing our approach 
is practically realizable, and that the combination of re-rank and zero-G feedbacks allows the users to train
the robot in few feedbacks. 

\header{Experiment setup:} In this study, five  users (not associated with this work) used our 
system to train Baxter for grocery checkout tasks, using 
zero-G and re-rank feedback. Zero-G was provided kinesthetically on the robot, while re-rank was elicited in a simulator (on a desktop computer).
A set of 10 tasks of varying difficulty level was presented to users one at a time, and they were 
instructed to provide feedback until they were satisfied with the top ranked trajectory. 
To quantify the quality of learning each user evaluated their own trajectories (self score), 
the trajectories learned of the other users (cross score), and those predicted by Oracle-svm, 
on a Likert scale of 1-5 (where 5 is the best). We also recorded the time a user took for each task---from 
start of training till the user was satisfied.
\begin{wraptable}[19]{r}{0.55\textwidth}
\centering
\begin{minipage}[t]{.55\textwidth}
\vspace*{0.3\captionReduceTop}
\vspace*{\captionReduceTop}
\caption{Shows learning statistics for each user averaged over all tasks. The number in parentheses is standard deviation.}
\vspace*{\captionReduceBot}
\resizebox{\linewidth}{!}{
\begin{tabular}{@{}cccccc}
\hline
\multirow{2}{*}{User}&\# Re-ranking & \# Zero-G & Average & \multicolumn{2}{c}{Trajectory Quality}\\
&feedback&feedback&time (min.)&self&cross\\\hline
1&5.4 (4.1)&3.3 (3.4)&7.8 (4.9)&3.8 (0.6)&4.0 (1.4)\\
2&1.8 (1.0)&1.7 (1.3)&4.6 (1.7)&4.3 (1.2)&3.6 (1.2)\\
3&2.9 (0.8)&2.0 (2.0)&5.0 (2.9)&4.4 (0.7)&3.2 (1.2)\\
4&3.2 (2.0)&1.5 (0.9)&5.3 (1.9)&3.0 (1.2)&3.7 (1.0)\\
5&3.6 (1.0)&1.9 (2.1)&5.0 (2.3)&3.5 (1.3)&3.3 (0.6)\\\hline
\end{tabular}
}
\label{tab:userstudy}
\centering
\includegraphics[width=0.48\linewidth]{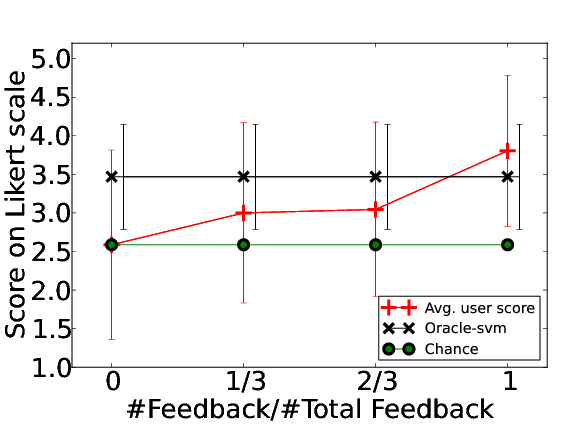}
\includegraphics[width=0.48\linewidth]{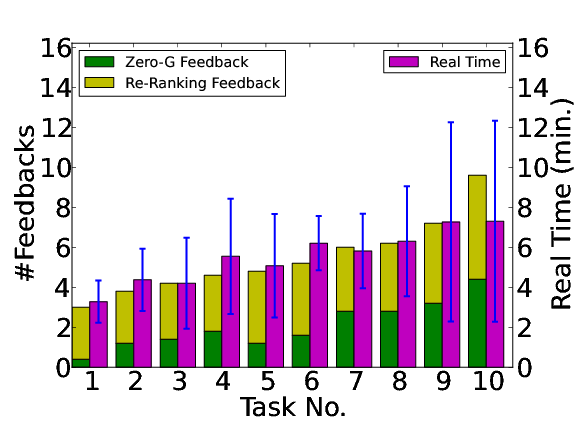}
\captionof{figure}{\textbf{(Left)} Average quality of the learned trajectory after every one-third of total feedback. 
\textbf{(Right)} Bar chart showing the average number of feedback and time required for 
each task. Task difficulty increases from 1 to 10.}
\label{fig:userplot}
\end{minipage}
\end{wraptable}
\header{Results from user study.} The study shows each user on an average took 3 re-rank and 2 zero-G 
feedbacks to train Baxter (Table~\ref{tab:userstudy}). 
Within 5 feedbacks the users were able to improve over Oracle-svm, Fig.~\ref{fig:userplot} (Left), 
consistent with our previous analysis.
Re-rank feedback was  popular for easier tasks, 
Fig.~\ref{fig:userplot} (Right). However as difficulty increased the users relied more 
on zero-G feedback, which allows  rectifying erroneous waypoints precisely.   
An average difference of 0.6 between users' self and cross score
suggests preferences marginally varied across the users. 

In terms of training time, 
each user took on average 5.5 minutes per-task, which we believe
is acceptable for most applications. Future research in human computer interaction, visualization 
and better user interface~\citep{Shneiderman03} could further reduce this time.
Despite its limited size, through user study we show our algorithm is
realizable in practice on high DoF manipulators. We hope this motivates researchers to build 
robotic systems capable of learning from \textit{non-expert} users.

For more details and video, please visit: \url{http://pr.cs.cornell.edu/coactive}

\vspace*{\sectionReduceTop}
\section{Conclusion} 
\vspace*{\sectionReduceBot}
\label{sec:conclusion}
In this paper we presented a co-active learning framework for training robots to select trajectories that 
obey a user's preferences. Unlike in standard learning from demonstration approaches, our framework 
does not require the user to provide optimal trajectories as training data, but can learn from iterative improvements. Despite only requiring weak feedback, our TPP learning algorithm has provable regret bounds and 
empirically performs well. In particular, we propose a set of trajectory features for which the TPP 
generalizes well on tasks which the robot has not seen before. In addition to the batch experiments, 
robotic experiments confirmed that incremental feedback generation is indeed feasible and that it 
leads to good learning results already after only a few iterations.

\textbf{Acknowledgments.} We thank Shikhar Sharma for help with the experiments. This research was supported by ARO, Microsoft Faculty fellowship and NSF Career award (to Saxena). 

\setlength{\bibsep}{0pt}
{\small
\bibliographystyle{plainnat}

\begin{thebibliography}{39}
\providecommand{\natexlab}[1]{#1}
\providecommand{\url}[1]{\texttt{#1}}
\expandafter\ifx\csname urlstyle\endcsname\relax
  \providecommand{\doi}[1]{doi: #1}\else
  \providecommand{\doi}{doi: \begingroup \urlstyle{rm}\Url}\fi

\bibitem[Abbeel et~al.(2010)Abbeel, Coates, and Ng]{Abbeel10}
P.~Abbeel, A.~Coates, and A.~Y. Ng.
\newblock Autonomous helicopter aerobatics through apprenticeship learning.
\newblock \emph{IJRR}, 29\penalty0 (13), 2010.

\bibitem[Akgun et~al.(2012)Akgun, Cakmak, Jiang, and Thomaz]{Akgun12}
B.~Akgun, M.~Cakmak, K.~Jiang, and A.~L. Thomaz.
\newblock Keyframe-based learning from demonstration.
\newblock \emph{IJSR}, 4\penalty0 (4):\penalty0 343--355, 2012.

\bibitem[Alterovitz et~al.(2007)Alterovitz, Sim{\'e}on, and
  Goldberg]{Alterovitz07}
R.~Alterovitz, T.~Sim{\'e}on, and K.~Goldberg.
\newblock The stochastic motion roadmap: A sampling framework for planning with
  markov motion uncertainty.
\newblock In \emph{RSS}, 2007.

\bibitem[Berg et~al.(2010)Berg, Abbeel, and Goldberg]{Berg10}
J.~V.~D. Berg, P.~Abbeel, and K.~Goldberg.
\newblock Lqg-mp: Optimized path planning for robots with motion uncertainty
  and imperfect state information.
\newblock In \emph{RSS}, 2010.

\bibitem[Calinon et~al.(2007)Calinon, Guenter, and Billard]{Calinon07}
S.~Calinon, F.~Guenter, and A.~Billard.
\newblock On learning, representing, and generalizing a task in a humanoid
  robot.
\newblock \emph{IEEE Transactions on Systems, Man, and Cybernetics}, 2007.

\bibitem[Dey et~al.(2012)Dey, Liu, Hebert, and Bagnell]{Dey12}
D.~Dey, T.~Y. Liu, M.~Hebert, and J.~A. Bagnell.
\newblock Contextual sequence prediction with application to control library
  optimization.
\newblock In \emph{RSS}, 2012.

\bibitem[Diankov(2010)]{Diankov10}
R.~Diankov.
\newblock \emph{Automated Construction of Robotic Manipulation Programs}.
\newblock PhD thesis, CMU, RI, 2010.

\bibitem[Dragan and Srinivasa(2013)]{Dragan13b}
A.~Dragan and S.~Srinivasa.
\newblock Generating legible motion.
\newblock In \emph{RSS}, 2013.

\bibitem[Green and Kelly(2007)]{Green07}
C.~J. Green and A.~Kelly.
\newblock Toward optimal sampling in the space of paths.
\newblock In \emph{ISRR}. 2007.

\bibitem[Jiang et~al.(2012{\natexlab{a}})Jiang, Lim, and
  Saxena]{jiang2012humancontext}
Y.~Jiang, M.~Lim, and A.~Saxena.
\newblock Learning object arrangements in 3d scenes using human context.
\newblock In \emph{ICML}, 2012{\natexlab{a}}.

\bibitem[Jiang et~al.(2012{\natexlab{b}})Jiang, Lim, Zheng, and
  Saxena]{jiang2012placingobjects}
Y.~Jiang, M.~Lim, C.~Zheng, and A.~Saxena.
\newblock Learning to place new objects in a scene.
\newblock \emph{IJRR}, 31\penalty0 (9), 2012{\natexlab{b}}.

\bibitem[Jiang et~al.(2013)Jiang, Koppula, and
  Saxena]{jiang-hallucinatinghumans-labeling3dscenes-cvpr2013}
Y.~Jiang, H.~Koppula, and A.~Saxena.
\newblock Hallucinated humans as the hidden context for labeling 3d scenes.
\newblock In \emph{CVPR}, 2013.

\bibitem[Joachims(2006)]{Joachims06}
T.~Joachims.
\newblock Training linear svms in linear time.
\newblock In \emph{KDD}, 2006.

\bibitem[Joachims et~al.(2009)Joachims, Finley, and Yu]{Joachims09}
T.~Joachims, T.~Finley, and C.~Yu.
\newblock Cutting-plane training of structural svms.
\newblock \emph{Mach Learn}, 77\penalty0 (1), 2009.

\bibitem[Karaman and Frazzoli(2010)]{Karaman10}
S.~Karaman and E.~Frazzoli.
\newblock Incremental sampling-based algorithms for optimal motion planning.
\newblock In \emph{RSS}, 2010.

\bibitem[Klingbeil et~al.(2011)Klingbeil, Rao, Carpenter, Ganapathi, Ng, and
  Khatib]{Klingbeil11}
E.~Klingbeil, D.~Rao, B.~Carpenter, V.~Ganapathi, A.~Y. Ng, and O.~Khatib.
\newblock Grasping with application to an autonomous checkout robot.
\newblock In \emph{ICRA}, 2011.

\bibitem[Kober and Peters(2011)]{Kober11}
J.~Kober and J.~Peters.
\newblock Policy search for motor primitives in robotics.
\newblock \emph{Machine Learning}, 84\penalty0 (1), 2011.

\bibitem[Koppula and Saxena(2013)]{Koppula13}
H.~S. Koppula and A.~Saxena.
\newblock Anticipating human activities using object affordances for reactive
  robotic response.
\newblock In \emph{RSS}, 2013.

\bibitem[Koppula et~al.(2011)Koppula, Anand, Joachims, and Saxena]{Koppula11}
H.~S. Koppula, A.~Anand, T.~Joachims, and A.~Saxena.
\newblock Semantic labeling of 3d point clouds for indoor scenes.
\newblock In \emph{NIPS}, 2011.

\bibitem[LaValle and Kuffner(2001)]{Lavalle01}
S.~M. LaValle and J.~J. Kuffner.
\newblock Randomized kinodynamic planning.
\newblock \emph{IJRR}, 20\penalty0 (5):\penalty0 378--400, 2001.

\bibitem[Lenz et~al.(2013)Lenz, Lee, and
  Saxena]{lenz2013_deeplearning_roboticgrasp}
I.~Lenz, H.~Lee, and A.~Saxena.
\newblock Deep learning for detecting robotic grasps.
\newblock In \emph{RSS}, 2013.

\bibitem[Levine and Koltun(2012)]{Levine12}
S.~Levine and V.~Koltun.
\newblock Continuous inverse optimal control with locally optimal examples.
\newblock In \emph{ICML}, 2012.

\bibitem[Mainprice et~al.(2011)Mainprice, Sisbot, Jaillet, Cort{\'e}s, Alami,
  and Sim{\'e}on]{Mainprice11}
J.~Mainprice, E.~A. Sisbot, L.~Jaillet, J.~Cort{\'e}s, R.~Alami, and
  T.~Sim{\'e}on.
\newblock Planning human-aware motions using a sampling-based costmap planner.
\newblock In \emph{ICRA}, 2011.

\bibitem[Manning et~al.(2008)Manning, Raghavan, and Sch{\"u}tze]{Manning08}
C.~D. Manning, P.~Raghavan, and H.~Sch{\"u}tze.
\newblock \emph{Introduction to information retrieval}, volume~1.
\newblock Cambridge University Press Cambridge, 2008.

\bibitem[Ratliff(2009)]{Ratliff09b}
N.~Ratliff.
\newblock \emph{Learning to Search: Structured Prediction Techniques for
  Imitation Learning}.
\newblock PhD thesis, CMU, RI, 2009.

\bibitem[Ratliff et~al.(2006)Ratliff, Bagnell, and Zinkevich]{Ratliff06}
N.~Ratliff, J.~A. Bagnell, and M.~Zinkevich.
\newblock Maximum margin planning.
\newblock In \emph{ICML}, 2006.

\bibitem[Ratliff et~al.(2007)Ratliff, Bradley, Bagnell, and
  Chestnutt]{Ratliff07b}
N.~Ratliff, D.~Bradley, J.~A. Bagnell, and J.~Chestnutt.
\newblock Boosting structured prediction for imitation learning.
\newblock In \emph{NIPS}, 2007.

\bibitem[Ratliff et~al.(2009{\natexlab{a}})Ratliff, Silver, and
  Bagnell]{Ratliff09a}
N.~Ratliff, D.~Silver, and J.~A. Bagnell.
\newblock Learning to search: Functional gradient techniques for imitation
  learning.
\newblock \emph{Autonomous Robots}, 27\penalty0 (1):\penalty0 25--53,
  2009{\natexlab{a}}.

\bibitem[Ratliff et~al.(2009{\natexlab{b}})Ratliff, Zucker, Bagnell, and
  Srinivasa]{Chomp}
N.~Ratliff, M.~Zucker, J.~A. Bagnell, and S.~Srinivasa.
\newblock Chomp: Gradient optimization techniques for efficient motion
  planning.
\newblock In \emph{ICRA}, 2009{\natexlab{b}}.

\bibitem[Saxena et~al.(2008)Saxena, Driemeyer, and
  Ng]{saxena2008roboticgrasping}
A.~Saxena, J.~Driemeyer, and A.Y. Ng.
\newblock Robotic grasping of novel objects using vision.
\newblock \emph{IJRR}, 27\penalty0 (2), 2008.

\bibitem[Shivaswamy and Joachims(2012)]{Shivaswamy12}
P.~Shivaswamy and T.~Joachims.
\newblock Online structured prediction via coactive learning.
\newblock In \emph{ICML}, 2012.

\bibitem[Shneiderman and Plaisant(2010)]{Shneiderman03}
B.~Shneiderman and C.~Plaisant.
\newblock \emph{Designing The User Interface: Strategies for Effective
  Human-Computer Interaction}.
\newblock Addison-Wesley Publication, 2010.

\bibitem[Sisbot et~al.(2007{\natexlab{a}})Sisbot, Marin, and Alami]{Sisbot07b}
E.~A. Sisbot, L.~F. Marin, and R.~Alami.
\newblock Spatial reasoning for human robot interaction.
\newblock In \emph{IROS}, 2007{\natexlab{a}}.

\bibitem[Sisbot et~al.(2007{\natexlab{b}})Sisbot, Marin-Urias, Alami, and
  Simeon]{Sisbot07}
E.~A. Sisbot, L.~F. Marin-Urias, R.~Alami, and T.~Simeon.
\newblock A human aware mobile robot motion planner.
\newblock \emph{IEEE Transactions on Robotics}, 2007{\natexlab{b}}.

\bibitem[Sucan et~al.(2012)Sucan, Moll, and Kavraki]{Ompl}
I.~A. Sucan, M.~Moll, and L.~E. Kavraki.
\newblock The {O}pen {M}otion {P}lanning {L}ibrary.
\newblock \emph{{IEEE} Robotics \& Automation Magazine}, 19\penalty0
  (4):\penalty0 72--82, 2012.
\newblock \url{http://ompl.kavrakilab.org}.

\bibitem[Vernaza and Bagnell(2012)]{Vernaza12}
P.~Vernaza and J.~A. Bagnell.
\newblock Efficient high dimensional maximum entropy modeling via symmetric
  partition functions.
\newblock In \emph{NIPS}, 2012.

\bibitem[Wilson et~al.(2012)Wilson, Fern, and Tadepalli]{Wilson12}
A.~Wilson, A.~Fern, and P.~Tadepalli.
\newblock A bayesian approach for policy learning from trajectory preference
  queries.
\newblock In \emph{NIPS}, 2012.

\bibitem[Zacharias et~al.(2011)Zacharias, Schlette, Schmidt, Borst, Rossmann,
  and Hirzinger]{Zacharias11}
F.~Zacharias, C.~Schlette, F.~Schmidt, C.~Borst, J.~Rossmann, and G.~Hirzinger.
\newblock Making planned paths look more human-like in humanoid robot
  manipulation planning.
\newblock In \emph{ICRA}, 2011.

\bibitem[Ziebart et~al.(2008)Ziebart, Maas, Bagnell, and Dey]{Ziebart08}
B.~D. Ziebart, A.~Maas, J.~A. Bagnell, and A.~K. Dey.
\newblock Maximum entropy inverse reinforcement learning.
\newblock In \emph{AAAI}, 2008.

\end{thebibliography}

}

\end{document}